\documentclass[9pt,twocolumn,format=sigconf, balance=true]{acmart}
\usepackage{libertine}
\usepackage{mathptmx}
\usepackage[T1]{fontenc}
\usepackage[utf8]{inputenc}
\usepackage[english]{babel}
\usepackage{refstyle}
\usepackage{standalone}
\usepackage{pgfplots}
\usepackage{pgffor}
\usepgfplotslibrary{groupplots}
\usepackage{todonotes}

\acmConference{SenSys}{November 2021}{Coimbra Portugal}{}
\setcopyright{none}
\renewcommand{\footnotetextcopyrightpermission}[1]{}
\newref{sec}{refcmd={Sec.~\ref{#1}}}
\newref{Sec}{refcmd={Sec.~\ref{#1}}}
\newref{Tab}{refcmd={Tab.~\ref{#1}}}
\newref{subsec}{refcmd={Sec.~\ref{#1}}}
\newref{Subsec}{refcmd={Sec.~\ref{#1}}}
\newref{fig}{refcmd={Fig.~\ref{#1}}}
\newref{Fig}{refcmd={Fig.~\ref{#1}}}

\begin{document}
\author{Tobias Kronauer}
\email{tobias.kronauer@bi-dd.de}
\affiliation{
	\institution{Barkhausen Institut}
	\city{Dresden}
	\country{Germany}
}
\author{Joshwa Pohlmann}
\email{joshwa.pohlmann@bi-dd.de}
\affiliation{
	\institution{Barkhausen Institut}
	\city{Dresden}
	\country{Germany}
}
\author{Maximilian Matth\'e}
\email{maximilian.matthe@bi-dd.de}
\affiliation{
	\institution{Barkhausen Institut}
	\city{Dresden}
	\country{Germany}
}
\author{Till Smejkal}
\email{till.smejkal@tu-dresden.de}
\affiliation{
	\institution{Operating Systems}
	\city{TU Dresden}
	\country{Germany}
}
\author{Gerhard Fettweis}
\email{gerhard.fettweis@bi-dd.de}
\affiliation{
	\institution{Barkhausen Institut}
	\city{Dresden}
	\country{Germany}
}
\title{Latency Analysis of ROS2 Multi-Node Systems}


\begin{abstract}
	The Robot Operating System 2 (ROS2) targets distributed real-time systems and is widely used in the robotics community. Especially in these systems, latency in data processing and communication can lead to instabilities. Though being highly configurable with respect to latency, ROS2 is often used with its default settings.
	
	In this paper, we investigate the end-to-end latency of ROS2 for distributed systems with default settings and different Data Distribution Service (DDS) middlewares. In addition, we profile the ROS2 stack and point out latency bottlenecks. Our findings indicate that end-to-end latency strongly depends on the used DDS middleware. Moreover, we show that ROS2 can lead to 50\,\% latency overhead compared to using low-level DDS communications. Our results imply guidelines for designing distributed ROS2 architectures and indicate possibilities for reducing the ROS2 overhead.
\end{abstract}

\maketitle

\section{Introduction\label{sec:Introduction}}

Through the advent of robotic applications, such as autonomous driving or household robotics, the robot operating system (\textit{ROS}) emerged as one of the most widely used software development frameworks. With more than tens of thousands of users~\cite{Metricsreport2019}, it is widely used in academia as well as in industry~\cite{RosRobots}.

Shortcomings of ROS1, but also its popularity led to the development of its successor, ROS2, which is developed by the Open Source Robotics Foundation (\textit{OSRF}) with many industrial contributors~\cite{Ros2ProjectGovernance, Casini2019}. Although being under heavy development, it is already used by notable companies~\cite{LgsvlWebpage} and the Open Source community~\cite{GazeboGitHubRepos}. Reasons for the development of ROS2 are new use cases, e.g. multiple robots, small embedded platforms, and real-time capability. Further, new technologies are available such as the Data Distribution Service (\textit{DDS})~\cite{DDSHomepage}.~\cite{WhyRos2}

ROS2 and its predecessor share the same core concept. Therefore, we refer to software architectures in ROS1 or ROS2 as \textit{ROS systems}. Among others, a ROS system comprises \textit{nodes}, \textit{messages}, and \textit{topics}. A \textit{node} is a software module performing computation. As ROS features the notion of modularity, a ROS system is usually composed of several nodes. These nodes exchange information by passing \textit{messages}. These messages only contain typed data structures that are standard primitive data types. A message is \textit{published} by a node to a given \textit{topic}, which is described by a string. Nodes that are interested in the data contained in this message \textit{subscribe} to that topic. Each node can have multiple subscribers and publishers.~\cite{Quigley2009}

\begin{figure}
	\centering
	\includegraphics[width=\linewidth]{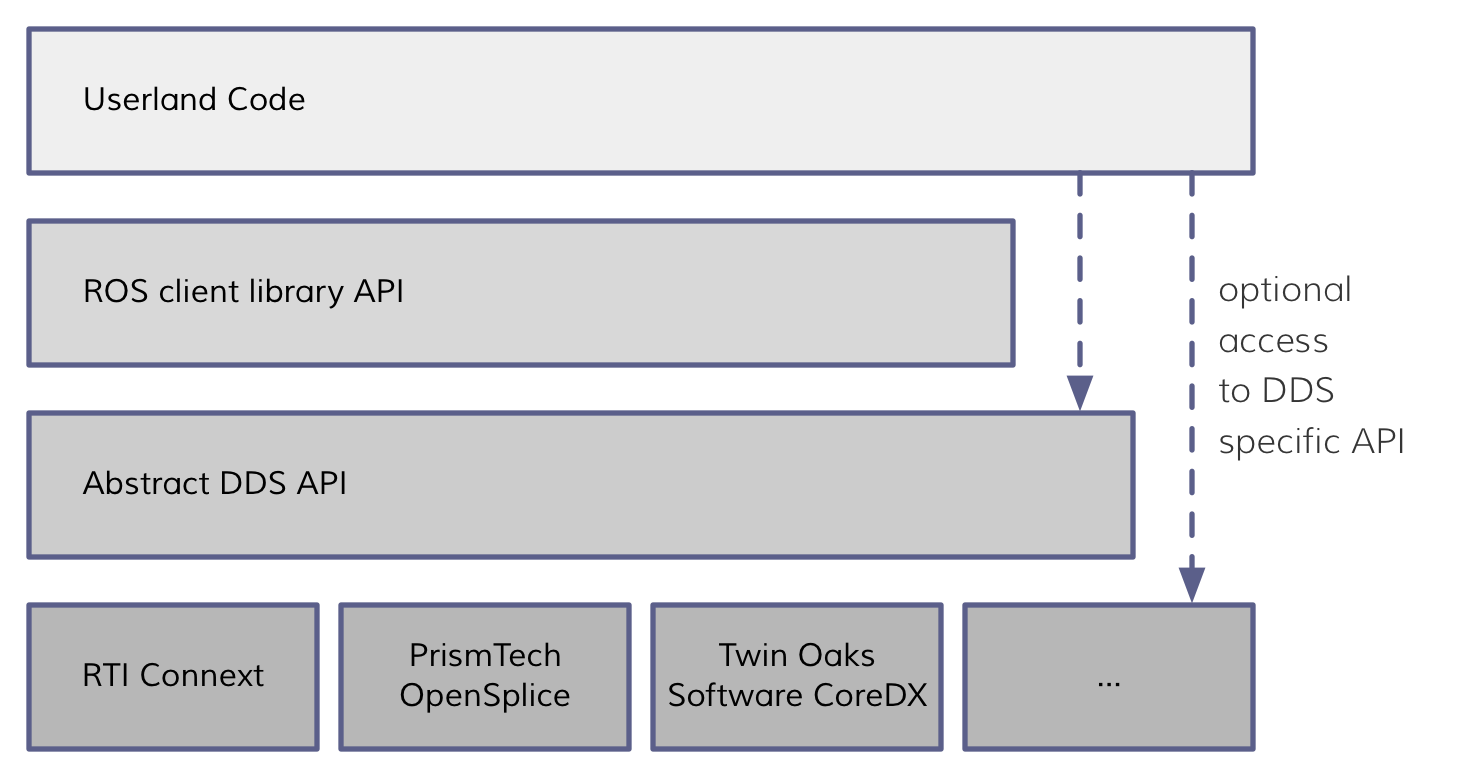}
	\Description{Between userland code and DDS middlware, ROS2 introduces two API layers.}
	\caption{API layout from ROS2, taken from \cite{ROSOnDDS}.}
	\label{fig:DDSAbstractionRos2}
\end{figure}

Although still using the publish/subscribe mechanism of ROS1, ROS2 builds its transport layer on a new middleware. The middleware is an implementation of the DDS standard~\cite{DDSHomepage} that is widely used for distributed, real-time systems. Without changing much of the usercode of ROS1, the goal was to hide the DDS middleware and its API to the ROS2 user as shown in \figref{DDSAbstractionRos2}. For that purpose, middleware interface modules (\texttt{rmw} in short for \textit{ros middleware}) were introduced. The most recent release of ROS2, Foxy Fitzroy, supports, among others, eProsima FastRTPS, Eclipse Cyclone DDS, and RTI Connext as DDS middleware.~\cite{ROSOnDDS, ROS2SupportedRMW}.

Since it is not the main focus of the paper, we will only briefly summarize DDS. The summary is based on the most recent DDS specification 1.4 \cite{DDSHomepage}. It describes a Data-Centric Publish-Subscribe model (\textit{DCPS}) for distributed applications enabling reliable, configurable, and real-time capable information between information points. Similar to ROS1 and ROS2, there are publishers that publish information of different data types. Each publisher has a \textit{DataWriter} that can be regarded as a source of information providing data of predefined type. The subscriber, on the other hand, is responsible for receiving the data sent by the publisher. A \textit{DataReader} is responsible for reading the information obtained by the subscriber. As in ROS, a topic fulfills the task to connect the publisher with one or more subscribers. A distinguishing feature of DDS is the use of \textit{quality of service} (QoS) policies. Each entity has an assigned QoS policy determining the behavior of each entity concerning communication and discovery. There is a huge variety of QoS parameters that can be set.

The described concept advocates the use of a distributed, modularized system. Applications that are based on a ROS system often use a software architecture of many nodes with well-defined interfaces~\cite{Naumann2018}. Running on \texttt{localhost}, shared memory and intraprocess communication may be used, significantly reducing latency. However, for distributed systems consisting of multiple hardware systems, this is not possible. In this work, we concentrate on the specifics provided by ROS \textit{by default}, i.e. that can be used by users without the necessity to tweak the configuration. We consider this a viable use case, as lots of potential users use the out-of-the-box behavior of ROS2.

Splitting the signal processing into several nodes, entails a certain latency to the system. Since real-time capability is one of the main feature claims of ROS2, the question arises how much latency is entailed by a system of many nodes compared to a system that performs the same computation in a single node. This leads to the central question of this paper: How does a ROS2 system cope with scalability? Latency plays a critical factor for new technologies, e.g. autonomous cars, edge cloud systems, and plenty of other applications enabled by 5G~\cite{Lema2017, Maheshwari2018, Voigtlaender2017}.

The paper is structured as follows: In \secref{RelatedWork}, we will summarize available publications that analyze ROS2 in terms of latency followed by the description of the contribution of our work. In the succeeding section, we will briefly explain our methodology for the latency evaluation. Results are presented in section \secref{Evaluation}. This work will be concluded with a future outlook in \secref{Conclusion}.

\section{Related Work\label{sec:RelatedWork}}

At the early stage of development, ROS2 was evaluated by~\cite{Maruyama2016} who used an alpha version and compared it with ROS1. Messages with sizes between 256\,B and 4\,MB were published at a 10\,Hz rate. The authors also vary the DDS middleware used by ROS2, i.e. RTI Connext, OpenSplice, and FastRTPS. The number of nodes that subscribe to a certain topic another node publishes to were varied as well. Latency, throughput via Ethernet, memory consumption, and threads per node were chosen as measurement metrics. Additionally, the overhead entailed by ROS2 to the latency is evaluated. They emphasize the importance of the message size for end-to-end latencies via Ethernet with a constant latency and ROS2 overhead until 64\,KB with the QoS policy being the major part. Further, they recommend a fragment size of 64\,KB that is used by the DDS middleware UDP protocol.

The results obtained were confirmed by~\cite{Reke2020} who evaluated ROS2 for autonomous cars on two KIA Niros with respect to real-time capability. They figured out that the results depend on the Linux kernel. Their software architecture is composed of 14 nodes on i7 Intel NUC. Messages were sent at a frequency of 33\,Hz and statistics of the time difference between two communicating nodes were calculated. By using a real-time kernel for Linux the jitter of the time difference could be significantly reduced. Using this information, the overall car behavior was analyzed under different scenarios.

The first official ROS2 release in December 2017, Ardent Apalone, was evaluated by~\cite{Gutierrez2018} using a real-time Linux kernel over Ethernet. By varying system loads, e.g. CPU load and concurrent traffic, round-trip latency was measured. One node was launched on a normal computer, another node was launched on an embedded device. A message was sent from the normal computer to the embedded device that sends the message back. The authors point out how this round-trip latency is influenced by the network and OS stack.

The previous papers emphasize the influence of various QoS parameters and DDS middlewares on the latency and throughput. \cite{Morita2018} propose an architecture that dynamically binds different DDS middlewares to ROS2 nodes to exploit performance benefits of the DDS middlewares under varying circumstances.

\cite{Wang2019} focused on improving the Inter Process Communication and compared it to the ROS2 implementation obtaining a lower latency and higher throughput. \cite{Kim2018} evaluated latency and throughput in combination with encryption.

\subsection{Contribution}

To the best of the author's knowledge, \cite{Maruyama2016} did the most comprehensive evaluation from a pure ROS2 perspective. Yet, the authors used a ROS2 alpha version and there have been a couple of releases in the meantime. Although \cite{Reke2020} did evaluations on realistic hardware, they only provided a brief, very high-level evaluation of their software architecture. The evaluations were mostly constrained to two or three nodes, i.e. it was not evaluated if the results can be scaled to a node system comprised of many nodes.

In this paper, we answer that question and provide the user with some guidelines to go along if the latency of a ROS system is to be decreased. For that, we perform an intra-layer profiling to concisely evaluate the bottlenecks in the communication in order to show possible implementation improvements for future ROS2 releases. Since distributed systems can be composed of different entities (robots, edge cloud, etc.), we define the use-case of reading a sensor in one component and post-processing it afterwards in other system components. Our parameters are then derived from this use case. We do not focus on the DDS middlewares as there are benchmarks on the homepage of the vendors. In addition, configuration strongly depends on the use case.

\section{Methodology\label{sec:Methodology}}

At first, we will define our use case in \secref{UseCaseDefinition}, which is followed by the description of the parameter space in \secref{ParameterSpace}. Afterwards, we will introduce measurement metrics and present our evaluation framework in \secref{MeasurementMetrics} and \secref{EvaluationFramework}.

\subsection{Use Case\label{sec:UseCaseDefinition}}

\begin{figure}[b]
	\centering
	\includegraphics[width=\linewidth]{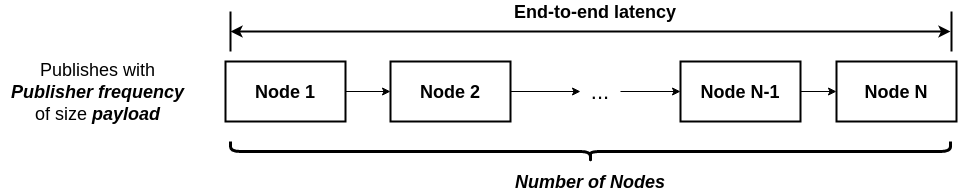}
	\Description{Introduced parameters are visualized.}
	\caption{Visualization of evaluation setup. Parameters are highlighted in italic.}
	\label{fig:EvaluationSetup}
\end{figure}

We focus on the use case of a distributed system. To generalize such system, we represent each system component (i.e. one computing node) by a single node performing all the processing. We assume that the processing of sensor readings and other additional calculations entail no latency, i.e. the message is sent out with the same frequency as the sampling rate of the sensor. We assume the communication pattern to be \textit{subsequent} as this is a \textit{worst-case} scenario leading to the highest latency between sending and receiving a message. This means, a first computing node is defined which sends out information to another component, which, upon receiving, sends this message to the third component, etc. The setup is depicted in \figref{EvaluationSetup}. We refer to a chain of nodes as \textit{data-processing pipeline}.

\subsection{Parameter Space\label{sec:ParameterSpace}}

Our parameters are derived from the use case of reading a sensor and post-processing it in a distributed fashion. The sampling frequency of the sensor is a variable parameter. Assuming that the sensor reading node only reads the sensor data and immediately publishes the message, this corresponds to the \textit{publisher frequency}. The data size of the sensor reading is variable as well. In the ROS system, the \textit{payload} of the message represents this quantity. The \textit{number of nodes}, which includes the start and end node in a data-processing pipeline can be modified as well. Moreover, we can change the Quality of Service settings. The aforementioned parameters remain constant throughout the data-processing pipeline.

\subsection{Measurement Metrics \label{sec:MeasurementMetrics}}

\textit{Latency} is our main measurement metric. In accordance with the previous papers mentioned in \secref{RelatedWork}, latency is defined between publishing a message in the first node and receiving the message in the last node, i.e. the call of the callback of the subscriber.

The focus of the evaluation is the latency entailed by the ROS system and not by the DDS middleware. Therefore, we want to \textit{profile} the call stack between publishing and subscriber callback for evaluating the overhead of ROS2 core and the middleware interfaces. As statistical quantity, we choose the 95\,\% percentile and do not analyze the distributions as such. This means 95\,\% of the calculated latencies are below this value. In general, the amount of messages that are below a certain threshold is of interest. As this depends on the actual use case, we decided to set this value to 95\,\%.

\subsection{Evaluation Framework \label{sec:EvaluationFramework}}

\begin{figure}[!]
	\centering
	\includegraphics[width=\linewidth]{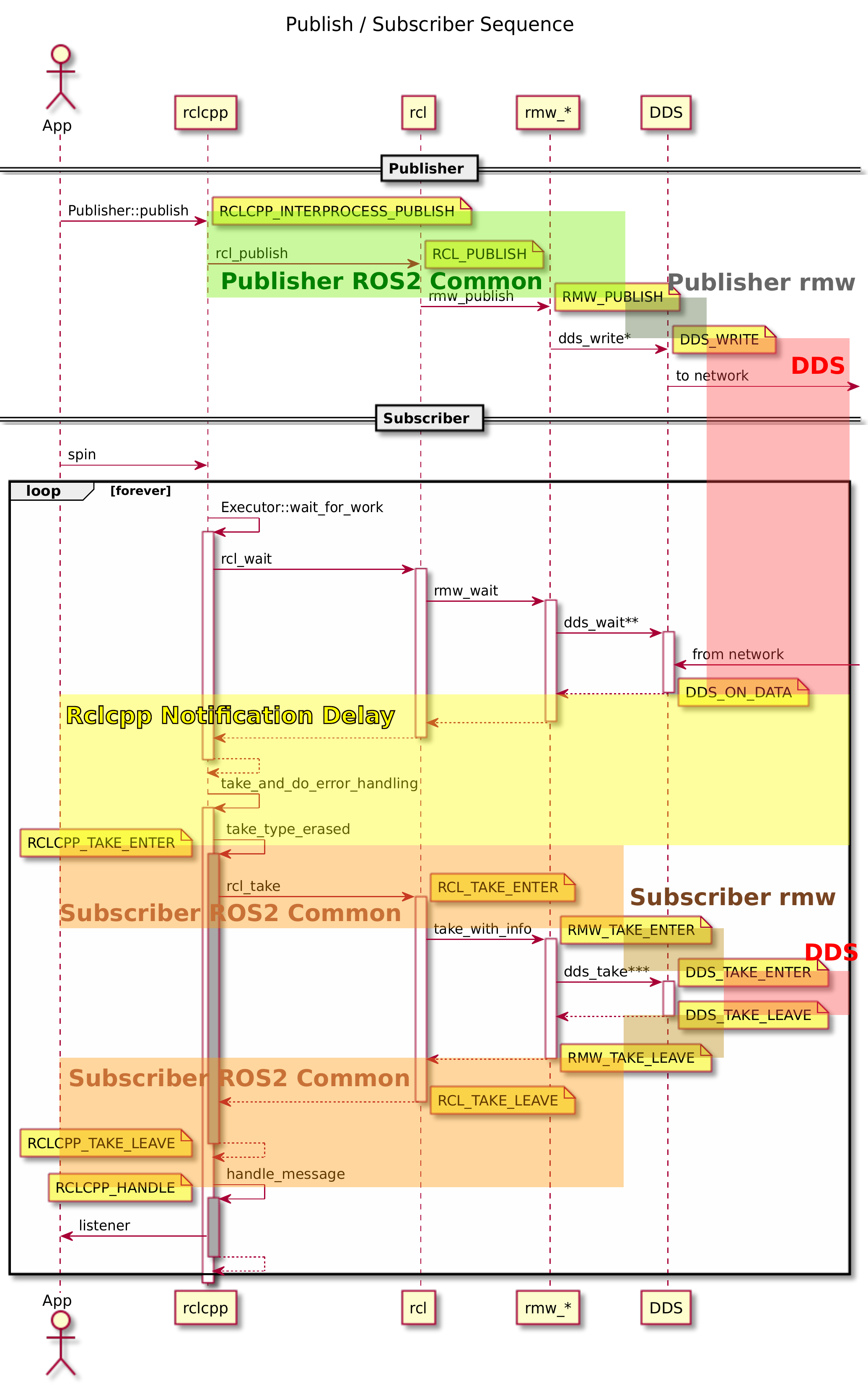}
	\Description{In a sequence diagram, we show required code adaptions for our profiling. Categorization is shown as well.}
	\caption{Sequence diagram of necessary adaptions to enable profiling. Notes indicate the layers. Colored boxes highlight the profiling categories.}
	\label{fig:SequenceDiagramProfiling}
\end{figure}

Relevant code to reproduce paper results can be found on GitHub~\cite{BIGithub}. \figref{SequenceDiagramProfiling} depicts the implementation of an intra-layer profiling for ROS2 messages. The implementation is bundled within a \texttt{docker} container containing the custom ROS2 build. The latency overhead of \texttt{docker} is negligible~\cite{Felter2015} provided that it is run in the \texttt{host} network. For the implementation, we added a \texttt{ros2profiling} package providing functions to generate timestamps since the Epoch in nanoseconds. It patches the ROS2 source tree to include timestamps into processed messages. Therefore, a minimum message size of 128\,B needs to be ensured. The timestamps correspond to the position of the notes in \figref{SequenceDiagramProfiling} in the ROS2 callstack. Timestamps are recorded when entering and leaving the method of the layer. Arrow annotations denote the called function and asterisks indicate placeholders for the actual function name, as this depends on the middleware. Timestamps were categorized into the following categories:
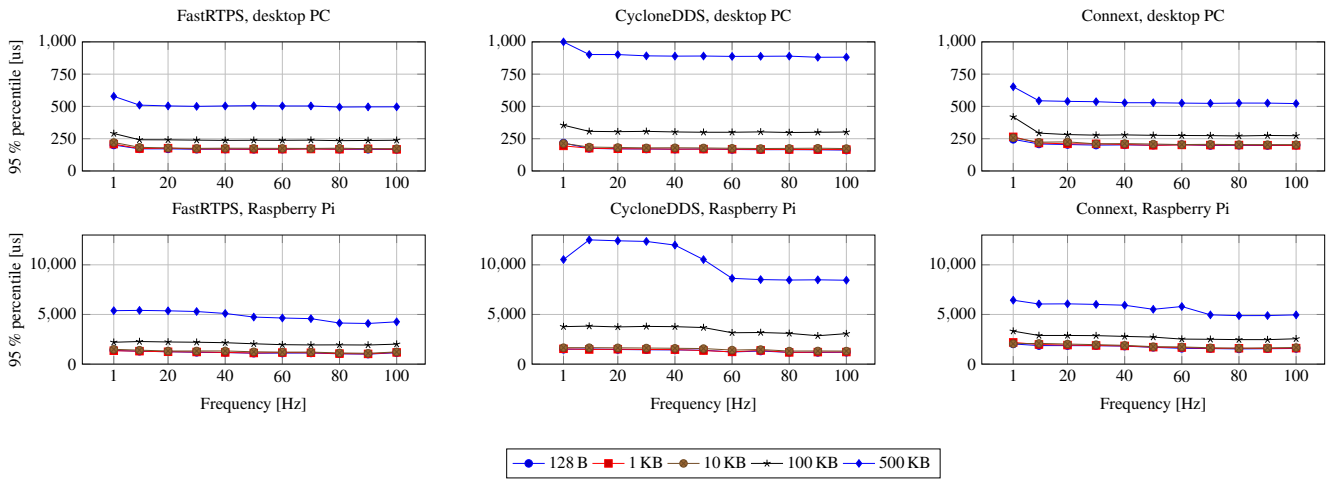
\begin{figure*}[!t]
	\resizebox{\linewidth}{!}{
			\pgfplotsset{scaled y ticks=false}
	\newcommand\statQuan{q95}
\begin{tikzpicture}
    \begin{groupplot}[
    			group style={
    					group size=3 by 2,
    					horizontal sep=2cm,
    					vertical sep=1.2cm,
    					ylabels at=edge left,
    					xlabels at=edge bottom,
   				},
	   			height=4cm,
   				width=8cm,
                ylabel={95\,\% percentile [us]}, 
                xlabel={Frequency [Hz]},
                legend entries={128\,B, 1\,KB, 10\,KB, 100\,KB, 500\,KB},
                legend style={at={(12,-5.2)}, anchor=north},
                legend columns=-1,
                legend to name=payloadLegend,
                grid=major,
                xmin=-10,
                xmax=110,
                xtick={1, 20, 40, 60, 80, 100},
                ymin=0]
        \nextgroupplot [title = {FastRTPS, desktop PC}, ytick={0, 250, 500, 750, 1000}, ymax=1000]	
        \foreach \MsgSize in {128b, 1kb, 10kb, 100kb, 500kb}
        {
        	\addplot table[x=Frequency, y=end2end_\statQuan, col sep=comma]{data/Workstation/frequency/fastrtps/fastrtps_3Nodes_1-10-20-30-40-50-60-70-80-90-100-Hz_\MsgSize_best-effort.csv};
        }
        \nextgroupplot [title = {CycloneDDS, desktop PC}, ytick={0, 250, 500, 750, 1000}, ymax=1000]	
		\foreach \MsgSize in {128b, 1kb, 10kb, 100kb, 500kb}
		{
			\addplot table[x=Frequency, y=end2end_\statQuan, col sep=comma]{data/Workstation/frequency/cyclonedds/cyclonedds_3Nodes_1-10-20-30-40-50-60-70-80-90-100-Hz_\MsgSize_best-effort.csv};
		}
        \nextgroupplot [title = {Connext, desktop PC}, ytick={0, 250, 500, 750, 1000}, ymax=1000]
		\foreach \MsgSize in {128b, 1kb, 10kb, 100kb, 500kb}
		{
			\addplot table[x=Frequency, y=end2end_\statQuan, col sep=comma]{data/Workstation/frequency/connext/connext_3Nodes_1-10-20-30-40-50-60-70-80-90-100-Hz_\MsgSize_best-effort.csv};
		}
     	\nextgroupplot [title = {FastRTPS, Raspberry Pi}, ymax=13000]
		\foreach \MsgSize in {128b, 1kb, 10kb, 100kb, 500kb}
		{
			\addplot table[x=Frequency, y=end2end_\statQuan, col sep=comma]{data/Raspi/frequency/fastrtps/fastrtps_3Nodes_1-10-20-30-40-50-60-70-80-90-100-Hz_\MsgSize_best-effort.csv};
		}
     	\nextgroupplot [title = {CycloneDDS, Raspberry Pi}, ymax=13000]
		\foreach \MsgSize in {128b, 1kb, 10kb, 100kb, 500kb}
		{
			\addplot table[x=Frequency, y=end2end_\statQuan, col sep=comma]{data/Raspi/frequency/cyclonedds/cyclonedds_3Nodes_1-10-20-30-40-50-60-70-80-90-100-Hz_\MsgSize_best-effort.csv};
		}
     	\nextgroupplot [title = {Connext, Raspberry Pi}, ymax=13000]
		\foreach \MsgSize in {128b, 1kb, 10kb, 100kb, 500kb}
		{
			\addplot table[x=Frequency, y=end2end_\statQuan, col sep=comma]{data/Raspi/frequency/connext/connext_3Nodes_1-10-20-30-40-50-60-70-80-90-100-Hz_\MsgSize_best-effort.csv};
		}

    \end{groupplot}
    \ref{payloadLegend}
\end{tikzpicture}
	}
	\Description{We evaluate the influence of the publisher frequency on the latency. Generally speaking, latency decreases with publisher frequency and increases with message size.}
	\caption{Investigating the influence of publisher frequency on latency with three nodes, QoS reliability is set to \texttt{BEST\_EFFORT}.}
	\label{fig:InfluencePublisherFrequency}
\end{figure*}
\begin{itemize}
	\item \textbf{DDS}: This category only contains latency entailed by the DDS transport via network and by the function call to actually receive the message.
	\item \textbf{Subscriber \texttt{rmw}} and \textbf{Publisher \texttt{rmw}}: Latency attributed to the \texttt{rmw} layer. Middleware-specific conversions from ROS2 messages to DDS messages that might require DDS utility functions but are not used for the transport of the message itself are contained in this category.
	\item \textbf{Publisher} and \textbf{Subscriber ROS2 Common}: Overhead entailed by ROS2 that is independent of the middleware.
	\item \textbf{Benchmarking}: This latency is entailed by code used for benchmarking the latency.
	\item \textbf{Rclcpp Notification Delay}: The time difference between the notification of DDS to ROS2 that new data is available and triggering of its actual retrieval.
\end{itemize}

As we are interested in the ROS2 overhead and not on the over-the-network performance, which is dictated by DDS and the network itself, latency is measured on one machine. In addition, this approach simplifies evaluation and test setup. Nodes of the data-processing pipeline are created in different processes. Note that this is contrast to our use case.

\section{Evaluation\label{sec:Evaluation}}

\begin{table}[b]
	\centering
	\caption{Variable parameter values.}
	\begin{tabular}{c|c}
		\hline
		\textbf{Publisher frequency}			& 1, 10, \dots, 90, 100 \\ \hline
		\textbf{Payload}                       & 128\,B, 1\,KB, 10\,KB, 100\,KB, 500\,KB \\ \hline
		\textbf{Number of Nodes}               & 3, 5, \dots, 21, 23 \\ \hline
		\textbf{DDS Backend}                   & Connext, FastRTPS, CycloneDDS \\ \hline
		\textbf{Reliability}                   & reliable, best effort \\ \hline
	\end{tabular}
	\label{tab:VariableParameterValues}
\end{table}

We compare two different hardware settings, namely a desktop PC, with an Intel i7-8700 CPU @ 3.2\,GHz x 6 and 32\,GB RAM. In addition, we use a Raspberry Pi 4 Model B Rev 1.1 with 4\,GB RAM. Kernel versions are 5.4.0-42-generic for the desktop PC and 5.4.0-1015-raspi for Raspberry Pi. The latest release of ROS2 at the time of writing (Foxy Fitzroy 20201211) is used as ROS2 version for our evaluation. We set the network device as \texttt{localhost} with UDP as transport layer. FastRTPS version is 2.0.1, CycloneDDS is 0.6.0, and Connext was updated to 6.0.1 as major improvements were made during this update.

The variable parameters are listed in \tabref{VariableParameterValues}. We assume a publisher frequency of maximum 100\,Hz, which is realistic for e.g. time-of-flight sensors. GPS sensors operate at an update rate of 1\,Hz up to 10\,Hz. We assume an update rate for cameras up to 100\,Hz. Therefore, the publisher frequency is in a range from 1\,Hz to 100\,Hz with a step size of 10\,Hz. As variable QoS policies, we modify the reliability from \texttt{RELIABLE} to \texttt{BEST\_EFFORT}, since they are frequently used. The number of nodes in the data-processing pipeline varies between three and 23 nodes with a step size of two. These values are arbitrarily chosen.

\begin{figure*}[t]
	\resizebox{\linewidth}{!}{
		\pgfplotsset{scaled y ticks=false}
\newcommand{\Frequencies}{1, 20, 50, 80, 100}
\newcommand{\statQuan}{q95}

\begin{tikzpicture}
    \begin{groupplot}[group style={
            group size={3 by 4},
            xlabels at=edge bottom,
            ylabels at=edge left,
            horizontal sep=1.5cm,
            vertical sep=1.2cm
        },
    	height=4cm,
    	width=8cm,
        legend entries={1\,Hz, 20\,Hz, 50\,Hz, 80\,Hz, 100\,Hz},
        legend style={at={(11.2,-11.8)}, anchor=north},
        legend columns=-1,
        legend to name=frequencyLegend,
        xlabel=Nodes,
        ylabel={95\,\% percentile [us]},
        xmin=0, xmax=25,
        xtick={3, 7, 11, 15, 19, 23},
        grid = major]
    \nextgroupplot[title={FastRTPS, 128\,B}, ymin=0, ymax=4000]
    \foreach \F in \Frequencies
    {
       	\addplot table[x=Nodes, y=end2end_\statQuan, col sep=comma]{data/Workstation/scalability/fastrtps/fastrtps_3-23Nodes_\F Hz_128b_best-effort.csv};
    }
   	\nextgroupplot[title={CyloneDDS, 128\,B}, ymin=0, ymax=4000]
	\foreach \F in \Frequencies
	{
       	\addplot table[x=Nodes, y=end2end_\statQuan, col sep=comma]{data/Workstation/scalability/cyclonedds/cyclonedds_3-23Nodes_\F Hz_128b_best-effort.csv};
	}
    \nextgroupplot[title={Connext, 128\,B}, ymin=0, ymax=4000]
	\foreach \F in \Frequencies
	{
       	\addplot table[x=Nodes, y=end2end_\statQuan, col sep=comma]{data/Workstation/scalability/connext/connext_3-23Nodes_\F Hz_128b_best-effort.csv};
	}
    \nextgroupplot[title={FastRTPS, 500\,KB}, ymin=0, ymax=12500]
    \foreach \F in \Frequencies
    {
       	\addplot table[x=Nodes, y=end2end_\statQuan, col sep=comma]{data/Workstation/scalability/fastrtps/fastrtps_3-23Nodes_\F Hz_500kb_best-effort.csv};
    }
    \nextgroupplot[title={CyloneDDS, 500\,KB}, ymin=0, ymax=12500]
    \foreach \F in \Frequencies
    {
       	\addplot table[x=Nodes, y=end2end_\statQuan, col sep=comma]{data/Workstation/scalability/cyclonedds/cyclonedds_3-23Nodes_\F Hz_500kb_best-effort.csv};
    }
	\nextgroupplot[title={Connext, 500\,KB}, ymin=0, ymax=12500]
	\foreach \F in \Frequencies
	{
       	\addplot table[x=Nodes, y=end2end_\statQuan, col sep=comma]{data/Workstation/scalability/connext/connext_3-23Nodes_\F Hz_500kb_best-effort.csv};
	}
    \nextgroupplot[title={FastRTPS, 128\,B}, ymin=0, ymax=20000]
	\foreach \F in \Frequencies
	{
		\addplot table[x=Nodes, y=end2end_\statQuan, col sep=comma]{data/Raspi/scalability/fastrtps/fastrtps_3-23Nodes_\F Hz_128b_best-effort.csv};
	}
	\nextgroupplot[title={CyloneDDS, 128\,B}, ymin=0, ymax=20000]
	\foreach \F in \Frequencies
	{
		\addplot table[x=Nodes, y=end2end_\statQuan, col sep=comma]{data/Raspi/scalability/cyclonedds/cyclonedds_3-23Nodes_\F Hz_128b_best-effort.csv};
	}
	\nextgroupplot[title={Connext, 128\,B}, ymin=0, ymax=20000]
	\foreach \F in \Frequencies
	{
		\addplot table[x=Nodes, y=end2end_\statQuan, col sep=comma]{data/Raspi/scalability/connext/connext_3-23Nodes_\F Hz_128b_best-effort.csv};
	}
    \nextgroupplot[title={FastRTPS, 100\,KB}, ymin=0, ymax=40000]
	\foreach \F in \Frequencies
	{
		\addplot table[x=Nodes, y=end2end_\statQuan, col sep=comma]{data/Raspi/scalability/fastrtps/fastrtps_3-23Nodes_\F Hz_100kb_best-effort.csv};
	}
	\nextgroupplot[title={CyloneDDS, 100\,KB}, ymin=0, ymax=40000]
	\foreach \F in \Frequencies
	{
		\addplot table[x=Nodes, y=end2end_\statQuan, col sep=comma]{data/Raspi/scalability/cyclonedds/cyclonedds_3-23Nodes_\F Hz_100kb_best-effort.csv};
	}
	\nextgroupplot[title={Connext, 100\,KB}, ymin=0, ymax=40000]
	\foreach \F in \Frequencies
	{
		\addplot table[x=Nodes, y=end2end_\statQuan, col sep=comma]{data/Raspi/scalability/connext/connext_3-23Nodes_\F Hz_100kb_best-effort.csv};
	}
	\end{groupplot}
    \ref{frequencyLegend}
\end{tikzpicture}
	}
	\Description{Scalability of a node system is evaluated for different hardwares and payloads. System behaves approximately linear.}
	\caption{Investigation of the scalability of a node system. We used payloads of 128\,B and of 500\,KB. The DDS middleware was varied. For visualization purposes, only a few frequencies were picked. Evaluation was performed on the desktop PC with QoS-reliability \texttt{BEST\_EFFORT}. Upper two rows: desktop PC, lower two rows: raspberry pi.}
	\label{fig:Scalability}
\end{figure*}
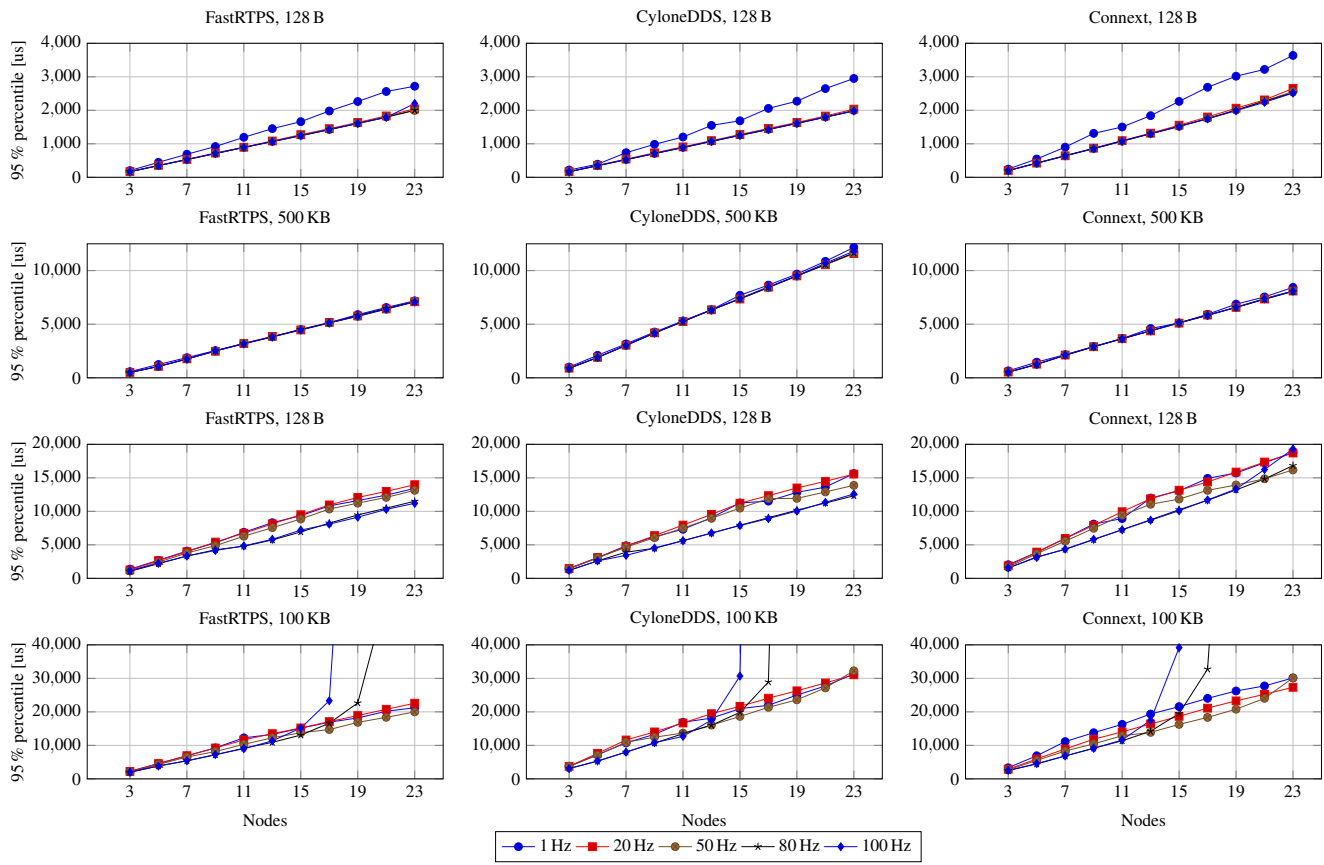
In all cases, we use a time duration for one configuration run of 60\,seconds. Depending on the publisher frequency, this results in a varying amount of samples, i.e. the higher the publisher frequency the higher the amount of samples.

The default settings of the Linux kernel are highly adaptive to the load and other external factors. In order to inhibit idle and energy saving mechanisms, which add additional noise to the latency measured, we change the kernel parameters: the \textit{scaling governor} is set to \texttt{userspace} and core frequencies are set to their maximum value. We deactivate the CPU idle by passing \texttt{cpuidle.off=1} as a kernel parameter. The option \texttt{no\_hz} disables the scheduler tick if no work is to be done on that CPU, which is why we deactivate it. For the desktop PC, we turn off hyperthreading.

\subsection{Evaluation of Publisher Frequency\label{sec:EvaluationPubF}}

We use three nodes with a publisher frequency between 1\,Hz and 100\,Hz as depicted in \figref{InfluencePublisherFrequency}. This corresponds to a ping-pong scenario. While sticking to the QoS reliability to \texttt{BEST\_EFFORT}, we use all possible payload sizes.

In the case of the desktop PC, for all DDS middlewares, we can see that latency decreases with increased frequency. This might be due to (unknown) energy saving features in other hardware devices or reprioritized thread scheduling. Further, we can see that the obtained latencies for 128\,B, 1\,KB, 10\,KB are equal, but increase for 100\,KB and 500\,KB. This can be explained by the maximum UDP packet size of 64\,KB which incurs fragmentation for large payloads. This additional overhead differs with the employed middleware. \mbox{FastRTPS} is slighty faster than Connext, whereas CycloneDDS entails the highest latency.

In the case of the Raspberry Pi, we can see that the latency decreases with the frequency as well, except for CycloneDDS with 1\,Hz. This can be due to low amount of samples. In addition, we can see a certain plateau between 10\,Hz and 40\,Hz for 500\,KB. In general, the latency obtained is significantly higher than for the desktop PC.

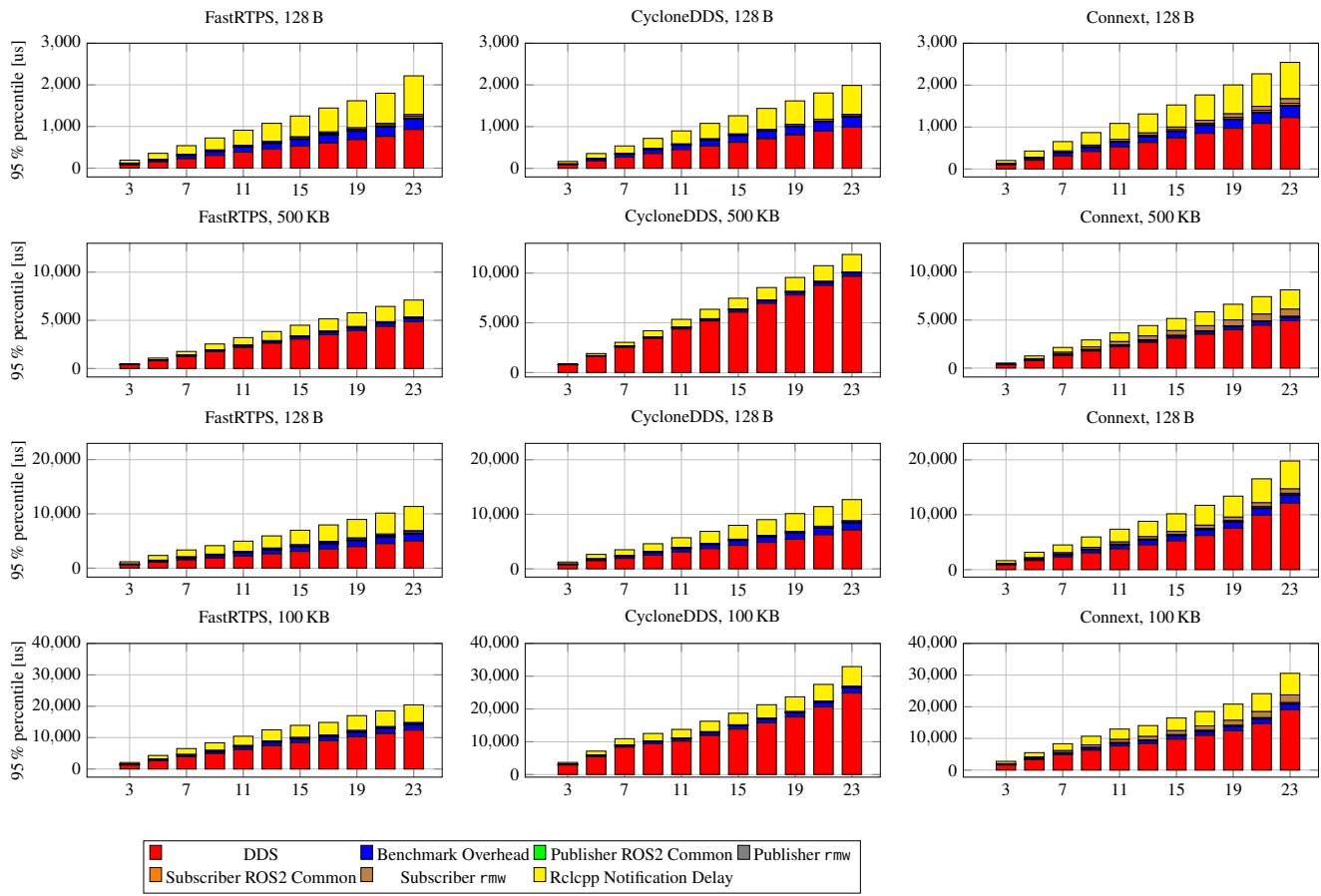
\begin{figure*}
	\resizebox{\linewidth}{!}{
		\pgfplotscreateplotcyclelist{profilingCycleList}{
	{black, fill=red},
	{black, fill=blue},
	{black, fill=green},
	{black, fill=gray},
	{black, fill=orange},
	{black, fill=brown},
	{black, fill=yellow},
}
\newcommand\statQuan{q95}
\pgfplotsset{scaled y ticks=false}
\begin{tikzpicture}
    \begin{groupplot}[
    	group style={
            group size={3 by 6},
            xlabels at=edge bottom,
            ylabels at=edge left,
            horizontal sep=1.5cm,
            vertical sep=1.2cm,
        },            
    	height=4cm,
    	width=8cm,
    	legend entries={DDS, Benchmark Overhead, Publisher ROS2 Common, Publisher \texttt{rmw}, Subscriber ROS2 Common, Subscriber \texttt{rmw}, Rclcpp Notification Delay},
    	legend columns=4,
    	legend style={at={(7.5, -12)}, anchor=north},
    	legend to name=profilingCategories,
        xlabel=Nodes,
        ylabel={95\,\% percentile [us]},
        grid = major,
        xmin=0, xmax=25,
        xtick={3,7,11,15,19,23}
    	]
   	\nextgroupplot[
   		title={FastRTPS, 128\,B}, 
   		ybar stacked, 
   		cycle list name=profilingCycleList,
   		ymax=3000
  	]
   	\foreach \Overhead in {DDS_\statQuan, benchmarkOverhead_\statQuan, Publisher_ROS2_Common_\statQuan, Publisher_rmw_\statQuan, Subscriber_ROS2_Common_\statQuan, Subscriber_rmw_\statQuan, Rclcpp_Notification_Delay_\statQuan}
   	{
   		\addplot+[ybar] table[x=Nodes, y={\Overhead}, col sep=comma]{data/Workstation/profiling/fastrtps_3-23Nodes_100Hz_128b_best-effort.csv};
   	}
    \nextgroupplot[
		title={CycloneDDS, 128\,B}, 
		ybar stacked, 
		cycle list name=profilingCycleList,
		ymax=3000
	]
   	\foreach \Overhead in {DDS_\statQuan, benchmarkOverhead_\statQuan, Publisher_ROS2_Common_\statQuan, Publisher_rmw_\statQuan, Subscriber_ROS2_Common_\statQuan, Subscriber_rmw_\statQuan, Rclcpp_Notification_Delay_\statQuan}
	{
	   	\addplot+[ybar] table[x=Nodes, y={\Overhead}, col sep=comma]{data/Workstation/profiling/cyclonedds_3-23Nodes_100Hz_128b_best-effort.csv};
	}    
	\nextgroupplot[
		title={Connext, 128\,B}, 
		ybar stacked, 
		cycle list name=profilingCycleList,
		ymax=3000
	]
   	\foreach \Overhead in {DDS_\statQuan, benchmarkOverhead_\statQuan, Publisher_ROS2_Common_\statQuan, Publisher_rmw_\statQuan, Subscriber_ROS2_Common_\statQuan, Subscriber_rmw_\statQuan, Rclcpp_Notification_Delay_\statQuan}
	{
		\addplot+[ybar] table[x=Nodes, y={\Overhead}, col sep=comma]{data/Workstation/profiling/connext_3-23Nodes_100Hz_128b_best-effort.csv};
	}
	\nextgroupplot[
		title={FastRTPS, 500\,KB}, 
		ybar stacked, 
		cycle list name=profilingCycleList,
		ymax=13000
	]
   	\foreach \Overhead in {DDS_\statQuan, benchmarkOverhead_\statQuan, Publisher_ROS2_Common_\statQuan, Publisher_rmw_\statQuan, Subscriber_ROS2_Common_\statQuan, Subscriber_rmw_\statQuan, Rclcpp_Notification_Delay_\statQuan}
	{
		\addplot+[ybar] table[x=Nodes, y={\Overhead}, col sep=comma]{data/Workstation/profiling/fastrtps_3-23Nodes_100Hz_500kb_best-effort.csv};
	}
	\nextgroupplot[
		title={CycloneDDS, 500\,KB}, 
		ybar stacked, 
		cycle list name=profilingCycleList,
		ymax=13000
	]
   	\foreach \Overhead in {DDS_\statQuan, benchmarkOverhead_\statQuan, Publisher_ROS2_Common_\statQuan, Publisher_rmw_\statQuan, Subscriber_ROS2_Common_\statQuan, Subscriber_rmw_\statQuan, Rclcpp_Notification_Delay_\statQuan}
	{
		\addplot+[ybar] table[x=Nodes, y={\Overhead}, col sep=comma]{data/Workstation/profiling/cyclonedds_3-23Nodes_100Hz_500kb_best-effort.csv};
	}
	\nextgroupplot[
		title={Connext, 500\,KB}, 
		ybar stacked, 
		cycle list name=profilingCycleList,
		ymax=13000
	]
   	\foreach \Overhead in {DDS_\statQuan, benchmarkOverhead_\statQuan, Publisher_ROS2_Common_\statQuan, Publisher_rmw_\statQuan, Subscriber_ROS2_Common_\statQuan, Subscriber_rmw_\statQuan, Rclcpp_Notification_Delay_\statQuan}
	{
		\addplot+[ybar] table[x=Nodes, y={\Overhead}, col sep=comma]{data/Workstation/profiling/connext_3-23Nodes_100Hz_500kb_best-effort.csv};
	}
   	\nextgroupplot[
		title={FastRTPS, 128\,B}, 
		ybar stacked, 
		cycle list name=profilingCycleList,
		ymax=23000
	]
	\foreach \Overhead in {DDS_\statQuan, benchmarkOverhead_\statQuan, Publisher_ROS2_Common_\statQuan, Publisher_rmw_\statQuan, Subscriber_ROS2_Common_\statQuan, Subscriber_rmw_\statQuan, Rclcpp_Notification_Delay_\statQuan}
	{
		\addplot+[ybar] table[x=Nodes, y={\Overhead}, col sep=comma]{data/Raspi/profiling/fastrtps_3-23Nodes_100Hz_128b_best-effort.csv};
	}
	\nextgroupplot[
		title={CycloneDDS, 128\,B}, 
		ybar stacked, 
		cycle list name=profilingCycleList,
		ymax=23000
	]
	\foreach \Overhead in {DDS_\statQuan, benchmarkOverhead_\statQuan, Publisher_ROS2_Common_\statQuan, Publisher_rmw_\statQuan, Subscriber_ROS2_Common_\statQuan, Subscriber_rmw_\statQuan, Rclcpp_Notification_Delay_\statQuan}
	{
		\addplot+[ybar] table[x=Nodes, y={\Overhead}, col sep=comma]{data/Raspi/profiling/cyclonedds_3-23Nodes_100Hz_128b_best-effort.csv};
	}    
	\nextgroupplot[
		title={Connext, 128\,B}, 
		ybar stacked, 
		cycle list name=profilingCycleList,
		ymax=23000
	]
	\foreach \Overhead in {DDS_\statQuan, benchmarkOverhead_\statQuan, Publisher_ROS2_Common_\statQuan, Publisher_rmw_\statQuan, Subscriber_ROS2_Common_\statQuan, Subscriber_rmw_\statQuan, Rclcpp_Notification_Delay_\statQuan}
	{
		\addplot+[ybar] table[x=Nodes, y={\Overhead}, col sep=comma]{data/Raspi/profiling/connext_3-23Nodes_100Hz_128b_best-effort.csv};
	}
	\nextgroupplot[
		title={FastRTPS, 100\,KB}, 
		ybar stacked, 
		cycle list name=profilingCycleList,
		ymax=40000
	]
	\foreach \Overhead in {DDS_\statQuan, benchmarkOverhead_\statQuan, Publisher_ROS2_Common_\statQuan, Publisher_rmw_\statQuan, Subscriber_ROS2_Common_\statQuan, Subscriber_rmw_\statQuan, Rclcpp_Notification_Delay_\statQuan}
	{
		\addplot+[ybar] table[x=Nodes, y={\Overhead}, col sep=comma]{data/Raspi/profiling/fastrtps_3-23Nodes_50Hz_100kb_best-effort.csv};
	}
	\nextgroupplot[
		title={CycloneDDS, 100\,KB}, 
		ybar stacked, 
		cycle list name=profilingCycleList,
		ymax=40000
	]
	\foreach \Overhead in {DDS_\statQuan, benchmarkOverhead_\statQuan, Publisher_ROS2_Common_\statQuan, Publisher_rmw_\statQuan, Subscriber_ROS2_Common_\statQuan, Subscriber_rmw_\statQuan, Rclcpp_Notification_Delay_\statQuan}
	{
		\addplot+[ybar] table[x=Nodes, y={\Overhead}, col sep=comma]{data/Raspi/profiling/cyclonedds_3-23Nodes_50Hz_100kb_best-effort.csv};
	}
	\nextgroupplot[
		title={Connext, 100\,KB}, 
		ybar stacked, 
		cycle list name=profilingCycleList,
		ymax=40000
	]
	\foreach \Overhead in {DDS_\statQuan, benchmarkOverhead_\statQuan, Publisher_ROS2_Common_\statQuan, Publisher_rmw_\statQuan, Subscriber_ROS2_Common_\statQuan, Subscriber_rmw_\statQuan, Rclcpp_Notification_Delay_\statQuan}
	{
		\addplot+[ybar] table[x=Nodes, y={\Overhead}, col sep=comma]{data/Raspi/profiling/connext_3-23Nodes_50Hz_100kb_best-effort.csv};
	}
  \end{groupplot}
\ref{profilingCategories}
\end{tikzpicture}
	}
	\Description{Results of profiling are shown. Most of the latency can be attributed to delay between notification and message retrieval for small message sizes. For larger message sizes, major latency contribution can be attributed to the DDS middleware.}
	\caption{Categorization of intra-process profiling of different latencies. Evaluation was performed with QoS reliability \texttt{BEST\_EFFORT}. Frequency is 100\,Hz for the desktop PC and message size 500\,KB. Concerning Raspberry Pi, frequency is 50\,Hz and 100\,KB. The latter one was fixed to these values, as a major amount of the sent messages was lost for higher payload and frequency.}
	\label{fig:Profiling}
\end{figure*}

\subsection{Evaluation of Scalability}

We evaluate the latency from starting to end node for the data-processing pipeline ranging from 3 to 23 nodes. As payload, we use 128\,B and 500\,KB. For visualization purposes, we restrict ourselves to a subset of the evaluated frequencies. Results are shown in \figref{Scalability}.

We can observe the same pattern as in \figref{InfluencePublisherFrequency}: the latency decreases if the frequency is increased. This effect is intensified with the length of the data-processing pipeline. As far as the desktop PC is concerned, for a payload of 128\,B, the results indicate a linear relationship between the number of nodes and latency. The outliers may occur due to too few samples, which is the case for lower frequencies. It needs to be further investigated if the relationship will be more linear if the number of samples is larger. 

Similar results were observed for the Raspberry Pi, whereas differences in latency with regard to frequency are more apparent. Although some linearity can be observed, more fluctuations occur. Interestingly, for Connext with 128\,B and a frequency of 100\,Hz, latency increases nonlinear for high node count. Regarding message size, we restricted ourselves to 100\,KB as plenty of messages (roughly 30\,\%) were lost for 500\,KB. In \figref{Scalability}, we can see for all middlewares the same pattern: for frequencies higher than 50\,Hz, latency increases extremely at 13 or 17 nodes. Hence, the raspberry pi seems to be overloaded in this area.

\subsection{Profiling}

The following findings apply for both hardwares: for a payload of 128\,B, the largest amount of latency can be attributed to the categories \textbf{DDS} and \textbf{Rclcpp Notification Delay} as seen in \figref{Profiling}. The overhead of ROS2 compared to raw DDS amounts up to 50\,\% for small messages. Using a payload of 500\,KB, one can clearly see that for all middlewares, the major part of the latency is due to the DDS middleware. As we focus on possible overhead reductions of ROS2 in this paper, we assume that the DDS middlewares cannot be changed. Thus, we can observe that major performance improvements can be obtained for the category \textbf{Rclcpp Notification Delay}. 

In addition, we can observe for the Raspberry Pi different conclusions for latency compared to the desktop PC: We can observe that Connext and CycloneDDS yield comparable results, whereas FastRTPS is significantly faster, which is different to the desktop PC. As for higher payload lots of messages were lost, we restricted ourselves to 100\,Kb and 50\,Hz. 

\subsection{Influence of QoS Reliability}

\begin{figure}[!]
	\resizebox{\linewidth}{!}{
		\newcommand\statQuan{q95}
\begin{tikzpicture}
    \begin{groupplot}[group style={
		group size={2 by 1},
		xlabels at=edge bottom,
		ylabels at=edge left,
		horizontal sep=1.5cm,
		vertical sep=1.2cm
	},
		height=4cm,
		width=8cm,
		legend entries={FastRTPS, CycloneDDS, Connext},
		legend style={at={(8.2,-1.2)}, anchor=north},
		legend columns=-1,
		legend to name=rmwLegend,
		xlabel=Nodes,
		ylabel={Relative Deviation [\%]},
		xmin=0, xmax=25,
		xtick={3, 7, 11, 15, 19, 23},
		ytick={-15, -7.5, 0, 7.5, 15}, ymin=-15, ymax=15,
		grid = major]
	\nextgroupplot[title={Desktop PC}]
        \addplot table[x=Nodes, y=end2end_\statQuan, col sep=comma]{data/Workstation/qos/relDeviationBeToRelQosfastrtps.csv};
        \addplot table[x=Nodes, y=end2end_\statQuan, col sep=comma]{data/Workstation/qos/relDeviationBeToRelQoscyclonedds.csv};
        \addplot table[x=Nodes, y=end2end_\statQuan, col sep=comma]{data/Workstation/qos/relDeviationBeToRelQosConnext.csv};
    \nextgroupplot[title={Raspberry Pi}]
        \addplot table[x=Nodes, y=end2end_\statQuan, col sep=comma]{data/Raspi/qos/relDeviationBeToRelQosfastrtps.csv};
	    \addplot table[x=Nodes, y=end2end_\statQuan, col sep=comma]{data/Raspi/qos/relDeviationBeToRelQoscyclonedds.csv};
	    \addplot table[x=Nodes, y=end2end_\statQuan, col sep=comma]{data/Raspi/qos/relDeviationBeToRelQosConnext.csv};
    \end{groupplot}
    \ref{rmwLegend}
\end{tikzpicture}
	}
	\Description{The relative difference in latency between \texttt{BEST\_EFFORT} and \texttt{RELIABLE} can reach up to 15\,\%.}
	\caption{Evaluation of influence of QoS reliability settings on latency. For the desktop PC, we choose 500\,KB and 100\,Hz. In the case of Raspberry Pi, we have a payload of 100\,KB and a publisher frequency of 50\,Hz.}
	\label{fig:QoS}
\end{figure}
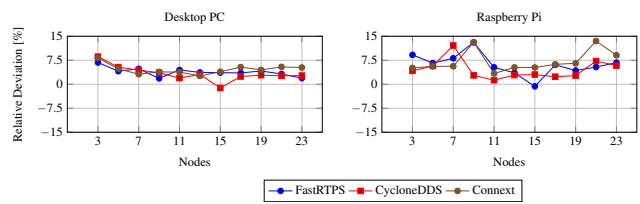

In the last sections, the QoS reliability policy was set to \texttt{BEST\_EFFORT} and kept as a constant parameter. We pick the highest possible throughput and calculate the relative deviation between the 95\,\% percentile of the latencies obtained with the QoS policy \texttt{BEST\_EFFORT} and \texttt{RELIABLE}. In our setup, using \texttt{RELIABLE} transmission imposed a latency overhead of up to 15\,\%, cf. \figref{QoS}.

\section{Conclusion and Future Outlook\label{sec:Conclusion}}

The goal of this paper was to provide the reader with simple guidelines if ROS2 is used for distributed systems. Given our parameter set, we discovered the following rules of thumb:

\begin{itemize}
	\item With a payload higher than the fragmentation size of UDP (here, 64\,KB), latency increases with the payload size.
	\item The higher the frequency, the lower the latency.
	\item Linearity with minor fluctuations for the 95\,\% percentile of latency with respect to node count could be observed.
	\item Latency critically depends on the hardware used and the parameter settings. No middleware yields the lowest latency in all cases.
	\item DDS middleware and the delay between message notification and message retrieval by ROS2 contribute the biggest portions to the overall latency.
	\item Latency is larger on Raspberry Pi and fluctuation in latency are more significant compared to the desktop PC. However, middlewares perform differently on the Raspberry Pi compared to the desktop PC.
\end{itemize}

During our evaluation, we observe that latency highly depends on energy saving features of the OS and the hardware. Our main focus was on the CPU. However, energy saving features of the NIC, e.g., might play an important role. This needs to be taken carefully into consideration if latency is to be mitigated for real-time critical applications. Further, a real-time kernel could be chosen.

In future work, one might consider an evaluation in a more realistic setup, i.e. with distributed systems. Network effects could be better evaluated. Aside from the 95\,percentile, one could verify if the latencies of the messages follow a certain distribution. The obtained information could then be used as an additional uncertainty for state estimation and incorporated into the Consensus Kalman Filter.

Although the amount of lost messages was negligible, differences between different DDS middlewares could be observed. This could be analyzed as well.

\begin{acks}
	This research was co-financed by public funding of the state of Saxony, Germany. We thank RTI for the license and support.
\end{acks}
\bibliographystyle{ACM-Reference-Format}
\bibliography{literature}


\begin{thebibliography}{23}


\ifx \showCODEN    \undefined \def \showCODEN     #1{\unskip}     \fi
\ifx \showDOI      \undefined \def \showDOI       #1{#1}\fi
\ifx \showISBNx    \undefined \def \showISBNx     #1{\unskip}     \fi
\ifx \showISBNxiii \undefined \def \showISBNxiii  #1{\unskip}     \fi
\ifx \showISSN     \undefined \def \showISSN      #1{\unskip}     \fi
\ifx \showLCCN     \undefined \def \showLCCN      #1{\unskip}     \fi
\ifx \shownote     \undefined \def \shownote      #1{#1}          \fi
\ifx \showarticletitle \undefined \def \showarticletitle #1{#1}   \fi
\ifx \showURL      \undefined \def \showURL       {\relax}        \fi
\providecommand\bibfield[2]{#2}
\providecommand\bibinfo[2]{#2}
\providecommand\natexlab[1]{#1}
\providecommand\showeprint[2][]{arXiv:#2}

\bibitem[\protect\citeauthoryear{??}{Gaz}{2020}]%
        {GazeboGitHubRepos}
 \bibinfo{year}{2020 (accessed August 17, 2020)}\natexlab{}.
\newblock \bibinfo{booktitle}{\emph{gazebo\_ros2\_control}}.
\newblock
\urldef\tempurl%
\url{https://github.com/gazebo_ros2_control}
\showURL{%
\tempurl}


\bibitem[\protect\citeauthoryear{Casini, Bla{\ss}, L{\"u}tkebohle, and
  Brandenburg}{Casini et~al\mbox{.}}{2019}]%
        {Casini2019}
\bibfield{author}{\bibinfo{person}{Daniel Casini}, \bibinfo{person}{Tobias
  Bla{\ss}}, \bibinfo{person}{Ingo L{\"u}tkebohle}, {and}
  \bibinfo{person}{Bj{\"o}rn~B. Brandenburg}.} \bibinfo{year}{2019}\natexlab{}.
\newblock \showarticletitle{{Response-Time Analysis of ROS 2 Processing Chains
  Under Reservation-Based Scheduling}}. In \bibinfo{booktitle}{\emph{31st
  Euromicro Conference on Real-Time Systems (ECRTS 2019)}}
  \emph{(\bibinfo{series}{Leibniz International Proceedings in Informatics
  (LIPIcs)}, Vol.~\bibinfo{volume}{133})},
  \bibfield{editor}{\bibinfo{person}{Sophie Quinton}} (Ed.).
  \bibinfo{publisher}{Schloss Dagstuhl--Leibniz-Zentrum fuer Informatik},
  \bibinfo{address}{Dagstuhl, Germany}, \bibinfo{pages}{6:1--6:23}.
\newblock
\showISBNx{978-3-95977-110-8}
\showISSN{1868-8969}
\urldef\tempurl%
\url{https://doi.org/10.4230/LIPIcs.ECRTS.2019.6}
\showDOI{\tempurl}


\bibitem[\protect\citeauthoryear{{Felter}, {Ferreira}, {Rajamony}, and
  {Rubio}}{{Felter} et~al\mbox{.}}{2015}]%
        {Felter2015}
\bibfield{author}{\bibinfo{person}{W. {Felter}}, \bibinfo{person}{A.
  {Ferreira}}, \bibinfo{person}{R. {Rajamony}}, {and} \bibinfo{person}{J.
  {Rubio}}.} \bibinfo{year}{2015}\natexlab{}.
\newblock \showarticletitle{An updated performance comparison of virtual
  machines and {L}inux containers}. In \bibinfo{booktitle}{\emph{2015 IEEE
  International Symposium on Performance Analysis of Systems and Software
  (ISPASS)}}. \bibinfo{pages}{171--172}.
\newblock


\bibitem[\protect\citeauthoryear{GitHub}{GitHub}{2021}]%
        {BIGithub}
\bibfield{author}{\bibinfo{person}{GitHub}.} \bibinfo{year}{2021 (accessed June
  06, 2021)}\natexlab{}.
\newblock \bibinfo{booktitle}{\emph{BI GitHub Projects}}.
\newblock
\urldef\tempurl%
\url{https://github.com/orgs/Barkhausen-Institut/projects}
\showURL{%
\tempurl}


\bibitem[\protect\citeauthoryear{Guti\'{e}rrez, Juan, Ugarte, and
  Vilches}{Guti\'{e}rrez et~al\mbox{.}}{2018}]%
        {Gutierrez2018}
\bibfield{author}{\bibinfo{person}{Carlos San~Vicente Guti\'{e}rrez},
  \bibinfo{person}{Lander Usategui~San Juan}, \bibinfo{person}{Irati~Zamalloa
  Ugarte}, {and} \bibinfo{person}{Víctor~Mayoral Vilches}.}
  \bibinfo{year}{2018}\natexlab{}.
\newblock \showarticletitle{Towards a distributed and real-time framework for
  robots: {Evaluation} of {ROS} 2.0 communications for real-time robotic
  applications}.
\newblock \bibinfo{journal}{\emph{arXiv:1809.02595 [cs]}}
  (\bibinfo{date}{Sept.} \bibinfo{year}{2018}).
\newblock
\urldef\tempurl%
\url{http://arxiv.org/abs/1809.02595}
\showURL{%
\tempurl}
\newblock
\shownote{arXiv: 1809.02595.}


\bibitem[\protect\citeauthoryear{Kim, Smereka, Cheung, Nepal, and Grobler}{Kim
  et~al\mbox{.}}{2018}]%
        {Kim2018}
\bibfield{author}{\bibinfo{person}{Jongkil Kim}, \bibinfo{person}{Jonathon~M.
  Smereka}, \bibinfo{person}{Calvin Cheung}, \bibinfo{person}{Surya Nepal},
  {and} \bibinfo{person}{Marthie Grobler}.} \bibinfo{year}{2018}\natexlab{}.
\newblock \showarticletitle{Security and {Performance} {Considerations} in
  {ROS} 2: {A} {Balancing} {Act}}.
\newblock \bibinfo{journal}{\emph{arXiv:1809.09566 [cs]}}
  (\bibinfo{date}{Sept.} \bibinfo{year}{2018}).
\newblock
\urldef\tempurl%
\url{http://arxiv.org/abs/1809.09566}
\showURL{%
\tempurl}


\bibitem[\protect\citeauthoryear{Lema, Laya, Mahmoodi, Cuevas, Sachs,
  Markendahl, and Dohler}{Lema et~al\mbox{.}}{2017}]%
        {Lema2017}
\bibfield{author}{\bibinfo{person}{Maria~A Lema}, \bibinfo{person}{Andres
  Laya}, \bibinfo{person}{Toktam Mahmoodi}, \bibinfo{person}{Maria Cuevas},
  \bibinfo{person}{Joachim Sachs}, \bibinfo{person}{Jan Markendahl}, {and}
  \bibinfo{person}{Mischa Dohler}.} \bibinfo{year}{2017}\natexlab{}.
\newblock \showarticletitle{{Business case and technology analysis for 5G low
  latency applications}}.
\newblock \bibinfo{journal}{\emph{IEEE Access}}  \bibinfo{volume}{5}
  (\bibinfo{year}{2017}), \bibinfo{pages}{5917--5935}.
\newblock


\bibitem[\protect\citeauthoryear{LGSVL}{LGSVL}{2020}]%
        {LgsvlWebpage}
\bibfield{author}{\bibinfo{person}{LGSVL}.} \bibinfo{year}{2020 (accessed
  August 17, 2020)}\natexlab{}.
\newblock \bibinfo{booktitle}{\emph{LGSVL Simulator}}.
\newblock
\urldef\tempurl%
\url{https://www.lgsvlsimulator.com/}
\showURL{%
\tempurl}


\bibitem[\protect\citeauthoryear{{Maheshwari}, {Raychaudhuri}, {Seskar}, and
  {Bronzino}}{{Maheshwari} et~al\mbox{.}}{2018}]%
        {Maheshwari2018}
\bibfield{author}{\bibinfo{person}{S. {Maheshwari}}, \bibinfo{person}{D.
  {Raychaudhuri}}, \bibinfo{person}{I. {Seskar}}, {and} \bibinfo{person}{F.
  {Bronzino}}.} \bibinfo{year}{2018}\natexlab{}.
\newblock \showarticletitle{{Scalability and Performance Evaluation of Edge
  Cloud Systems for Latency Constrained Applications}}. In
  \bibinfo{booktitle}{\emph{{2018 IEEE/ACM Symposium on Edge Computing
  (SEC)}}}. \bibinfo{pages}{286--299}.
\newblock


\bibitem[\protect\citeauthoryear{Maruyama, Kato, and Azumi}{Maruyama
  et~al\mbox{.}}{2016}]%
        {Maruyama2016}
\bibfield{author}{\bibinfo{person}{Yuya Maruyama}, \bibinfo{person}{Shinpei
  Kato}, {and} \bibinfo{person}{Takuya Azumi}.}
  \bibinfo{year}{2016}\natexlab{}.
\newblock \showarticletitle{Exploring the performance of {ROS2}}. In
  \bibinfo{booktitle}{\emph{Proceedings of the 13th {International}
  {Conference} on {Embedded} {Software} - {EMSOFT} '16}}.
  \bibinfo{publisher}{ACM Press}, \bibinfo{address}{Pittsburgh, Pennsylvania},
  \bibinfo{pages}{1--10}.
\newblock
\showISBNx{978-1-4503-4485-2}
\urldef\tempurl%
\url{https://doi.org/10.1145/2968478.2968502}
\showDOI{\tempurl}


\bibitem[\protect\citeauthoryear{Morita and Matsubara}{Morita and
  Matsubara}{2018}]%
        {Morita2018}
\bibfield{author}{\bibinfo{person}{Ren Morita} {and} \bibinfo{person}{Katsuya
  Matsubara}.} \bibinfo{year}{2018}\natexlab{}.
\newblock \showarticletitle{Dynamic {Binding} a {Proper} {DDS} {Implementation}
  for {Optimizing} {Inter}-{Node} {Communication} in {ROS2}}. In
  \bibinfo{booktitle}{\emph{2018 {IEEE} 24th {International} {Conference} on
  {Embedded} and {Real}-{Time} {Computing} {Systems} and {Applications}
  ({RTCSA})}}. \bibinfo{pages}{246--247}.
\newblock
\urldef\tempurl%
\url{https://doi.org/10.1109/RTCSA.2018.00043}
\showDOI{\tempurl}
\newblock
\shownote{ISSN: 2325-1301.}


\bibitem[\protect\citeauthoryear{{Naumann}, {Poggenhans}, {Lauer}, and
  {Stiller}}{{Naumann} et~al\mbox{.}}{2018}]%
        {Naumann2018}
\bibfield{author}{\bibinfo{person}{M. {Naumann}}, \bibinfo{person}{F.
  {Poggenhans}}, \bibinfo{person}{M. {Lauer}}, {and} \bibinfo{person}{C.
  {Stiller}}.} \bibinfo{year}{2018}\natexlab{}.
\newblock \showarticletitle{{CoInCar-Sim: An Open-Source Simulation Framework
  for Cooperatively Interacting Automobiles}}. In
  \bibinfo{booktitle}{\emph{{2018 IEEE Intelligent Vehicles Symposium (IV)}}}.
  \bibinfo{pages}{1--6}.
\newblock


\bibitem[\protect\citeauthoryear{OMG}{OMG}{2020}]%
        {DDSHomepage}
\bibfield{author}{\bibinfo{person}{OMG}.} \bibinfo{year}{2015 (accessed August
  17, 2020)}\natexlab{}.
\newblock \bibinfo{booktitle}{\emph{{Data Distribution Service}}}.
\newblock
\urldef\tempurl%
\url{https://www.omg.org/spec/DDS/}
\showURL{%
\tempurl}


\bibitem[\protect\citeauthoryear{OSRF}{OSRF}{2020b}]%
        {Metricsreport2019}
\bibfield{author}{\bibinfo{person}{OSRF}.} \bibinfo{year}{2019 (accessed August
  17, 2020)}\natexlab{b}.
\newblock \bibinfo{booktitle}{\emph{Community Metrics Report}}.
\newblock
\urldef\tempurl%
\url{http://download.ros.org/downloads/metrics/metrics-report-2019-07.pdf}
\showURL{%
\tempurl}


\bibitem[\protect\citeauthoryear{OSRF}{OSRF}{2020d}]%
        {ROSOnDDS}
\bibfield{author}{\bibinfo{person}{OSRF}.} \bibinfo{year}{2019 (accessed August
  17, 2020)}\natexlab{d}.
\newblock \bibinfo{booktitle}{\emph{ROS on DDS}}.
\newblock
\urldef\tempurl%
\url{https://design.ros2.org/articles/ros_on_dds.html}
\showURL{%
\tempurl}


\bibitem[\protect\citeauthoryear{OSRF}{OSRF}{2020a}]%
        {ROS2SupportedRMW}
\bibfield{author}{\bibinfo{person}{OSRF}.} \bibinfo{year}{2020 (accessed August
  17, 2020)}\natexlab{a}.
\newblock \bibinfo{booktitle}{\emph{About different ROS 2 DDS/RTPS vendors}}.
\newblock
\urldef\tempurl%
\url{https://index.ros.org/doc/ros2/Concepts/DDS-and-ROS-middleware-implementations/}
\showURL{%
\tempurl}


\bibitem[\protect\citeauthoryear{OSRF}{OSRF}{2020c}]%
        {Ros2ProjectGovernance}
\bibfield{author}{\bibinfo{person}{OSRF}.} \bibinfo{year}{2020 (accessed August
  17, 2020)}\natexlab{c}.
\newblock \bibinfo{booktitle}{\emph{Project Governance}}.
\newblock
\urldef\tempurl%
\url{https://index.ros.org/doc/ros2/Governance/#governance}
\showURL{%
\tempurl}


\bibitem[\protect\citeauthoryear{OSRF}{OSRF}{2020e}]%
        {RosRobots}
\bibfield{author}{\bibinfo{person}{OSRF}.} \bibinfo{year}{2020 (accessed August
  17, 2020)}\natexlab{e}.
\newblock \bibinfo{booktitle}{\emph{ROS Robots}}.
\newblock
\urldef\tempurl%
\url{https://robots.ros.org/}
\showURL{%
\tempurl}


\bibitem[\protect\citeauthoryear{OSRF}{OSRF}{2020f}]%
        {WhyRos2}
\bibfield{author}{\bibinfo{person}{OSRF}.} \bibinfo{year}{2020 (accessed August
  17, 2020)}\natexlab{f}.
\newblock \bibinfo{booktitle}{\emph{Why ROS 2?}}
\newblock
\urldef\tempurl%
\url{https://design.ros2.org/articles/why_ros2.html}
\showURL{%
\tempurl}


\bibitem[\protect\citeauthoryear{Quigley, Conley, Gerkey, Faust, Foote, Leibs,
  Wheeler, and Ng}{Quigley et~al\mbox{.}}{2009}]%
        {Quigley2009}
\bibfield{author}{\bibinfo{person}{Morgan Quigley}, \bibinfo{person}{Ken
  Conley}, \bibinfo{person}{Brian Gerkey}, \bibinfo{person}{Josh Faust},
  \bibinfo{person}{Tully Foote}, \bibinfo{person}{Jeremy Leibs},
  \bibinfo{person}{Rob Wheeler}, {and} \bibinfo{person}{Andrew~Y Ng}.}
  \bibinfo{year}{2009}\natexlab{}.
\newblock \showarticletitle{{ROS: an open-source Robot Operating System}}. In
  \bibinfo{booktitle}{\emph{ICRA workshop on open source software}},
  Vol.~\bibinfo{volume}{3}. Kobe, Japan, \bibinfo{pages}{5}.
\newblock


\bibitem[\protect\citeauthoryear{Reke, Peter, Schulte-Tigges, Schiffer,
  Ferrein, Walter, and Matheis}{Reke et~al\mbox{.}}{2020}]%
        {Reke2020}
\bibfield{author}{\bibinfo{person}{Michael Reke}, \bibinfo{person}{Daniel
  Peter}, \bibinfo{person}{Joschua Schulte-Tigges}, \bibinfo{person}{Stefan
  Schiffer}, \bibinfo{person}{Alexander Ferrein}, \bibinfo{person}{Thomas
  Walter}, {and} \bibinfo{person}{Dominik Matheis}.}
  \bibinfo{year}{2020}\natexlab{}.
\newblock \showarticletitle{A {Self}-{Driving} {Car} {Architecture} in {ROS2}}.
  In \bibinfo{booktitle}{\emph{2020 {International} {SAUPEC}/{RobMech}/{PRASA}
  {Conference}}}. \bibinfo{pages}{1--6}.
\newblock
\urldef\tempurl%
\url{https://doi.org/10.1109/SAUPEC/RobMech/PRASA48453.2020.9041020}
\showDOI{\tempurl}


\bibitem[\protect\citeauthoryear{{Voigtl\"{a}nder}, {Ramadan}, {Eichinger},
  {Lenz}, {Pensky}, and {Knoll}}{{Voigtl\"{a}nder} et~al\mbox{.}}{2017}]%
        {Voigtlaender2017}
\bibfield{author}{\bibinfo{person}{F. {Voigtl\"{a}nder}}, \bibinfo{person}{A.
  {Ramadan}}, \bibinfo{person}{J. {Eichinger}}, \bibinfo{person}{C. {Lenz}},
  \bibinfo{person}{D. {Pensky}}, {and} \bibinfo{person}{A. {Knoll}}.}
  \bibinfo{year}{2017}\natexlab{}.
\newblock \showarticletitle{{5G for Robotics: Ultra-Low Latency Control of
  Distributed Robotic Systems}}. In \bibinfo{booktitle}{\emph{2017
  International Symposium on Computer Science and Intelligent Controls
  (ISCSIC)}}. \bibinfo{pages}{69--72}.
\newblock


\bibitem[\protect\citeauthoryear{Wang, Tan, Hu, Manocha, and Hu}{Wang
  et~al\mbox{.}}{2019}]%
        {Wang2019}
\bibfield{author}{\bibinfo{person}{Yu-Ping Wang}, \bibinfo{person}{Wende Tan},
  \bibinfo{person}{Xu-Qiang Hu}, \bibinfo{person}{Dinesh Manocha}, {and}
  \bibinfo{person}{Shi-Min Hu}.} \bibinfo{year}{2019}\natexlab{}.
\newblock \showarticletitle{{TZC}: {Efficient} {Inter}-{Process}
  {Communication} for {Robotics} {Middleware} with {Partial} {Serialization}}.
  In \bibinfo{booktitle}{\emph{2019 {IEEE}/{RSJ} {International} {Conference}
  on {Intelligent} {Robots} and {Systems} ({IROS})}}.
  \bibinfo{pages}{7805--7812}.
\newblock
\urldef\tempurl%
\url{https://doi.org/10.1109/IROS40897.2019.8968462}
\showDOI{\tempurl}
\newblock
\shownote{ISSN: 2153-0866.}


\end{thebibliography}
\balance

\end{document}